\documentclass[
]{ceurart}

\usepackage{listings}
\begin{document}

\copyrightyear{2022}
\copyrightclause{Copyright for this paper by its authors.
  Use permitted under Creative Commons License Attribution 4.0
  International (CC BY 4.0).}

\conference{MACLEAN: MAChine Learning for EArth ObservatioN Workshop 2022,  in conjunction with ECML/PKDD (European Conference on Machine Learning and Principles and Practice of Knowledge Discovery in Databases) 2022, Grenoble, September 19, 2022}

\title{Towards unsupervised assessment with open-source data of the accuracy of deep learning-based distributed PV mapping}
\tnotemark[1]
\tnotetext[1]{Supported by RTE France.}

\author[1,2]{Gabriel Kasmi}[%
orcid=0000-0002-7774-4302,
email=firstname.lastname@minesparis.psl.eu,
url=https://gabrielkasmi.github.io/,
]
\cormark[1]

\author[2]{Laurent Dubus}[%
orcid=0000-0002-3987-646X,
]

\author[1]{Philippe Blanc}[%
orcid=0000-0002-6345-0004,
]

\author[1]{Yves-Marie Saint-Drenan}[%
orcid=0000-0003-1471-1092,
]

\cortext[1]{Corresponding author.}

\address[1]{MINES Paris, Université PSL, Centre Observation Impacts Energie (O.I.E.),
  06904 Sophia Antipolis, France}
\address[2]{RTE - Réseau de transport d'électricité,
92073 Paris La Défense, France}

\begin{abstract}
Photovoltaic (PV) energy is rapidly growing and key to mitigating the energy crisis. However, distributed PV generation, which amounts to half of the PV installed capacity, is typically unavailable to transmission system operators (TSOs), making it increasingly difficult to balance the load and supply and avoid grid congestions. To assess distributed PV generation, TSOs need precise knowledge regarding the metadata of distributed PV installations. Many remote sensing-based approaches have been proposed to map these installations in recent years. However, to use these methods in industrial processes, assessing their accuracy over the mapping area, i.e., the area covered by the model during deployment, is necessary. We define the downstream task accuracy (DTA) as the accuracy over the mapping area, automatically computed using publicly available data sources and the model's outputs and expressed in an interpretable way for operators. We benchmark existing models for distributed PV mapping and show how they perform in terms of DTA. We show that the accuracy computed on the test set overestimates by about 30 percentage points the accuracy on the mapping area. Our approach paves the way for safer integration of deep-learning-based pipelines for remote PV mapping. Code is available at \url{https://github.com/gabrielkasmi/dsfrance}.
\end{abstract}


\maketitle

\section{Introduction}

Integrating growing amounts of PV energy into the electric grid is challenging for transmission system operators (TSOs). Like other renewable energy sources (RES), PV energy is weather-dependent and highly decentralized. Therefore, to ensure its optimal integration on the grid, accurate measurements of the production and accurate production forecasts are necessary. Otherwise, increased PV penetration could result in increased congestion, imbalances, and supplementary reserve requests \cite{pierro2022progress}.

TSOs are responsible for the balance between the load and supply of electricity. They rely on production measurements and forecasts for all energy sources to ensure this balance. In the case of PV energy, only the production stemming from power plants is accessible to the TSO through real-time measurements (when their installed capacity is larger than 1 $MW_p$) or estimations that use regional PV models \cite{sengupta2021best}. The remaining installations, often referred to as distributed PV, are invisible to the system operator \cite{shaker2015data}. As distributed PV amounts to half the PV installed capacity, TSOs must derive methods to estimate the production of these invisible distributed PV installations.

A straightforward approach would be to expand regional PV models to distributed PV. These models estimate the PV power generation over a given area, taking a limited set of PV characteristics (installed capacity, tilt, and azimuth angles) and weather data \cite{saint2017probabilistic,sengupta2021best}. As there is no centralized record of the characteristics of distributed PV installations \cite{killinger2018search}, a popular solution is to map the characteristics of distributed PV installations using earth observation data and deep learning models. Characteristics usually include the localization, tilt, azimuth, and installed capacity of the distributed PV installations \cite{killinger2018search}. Works such as \cite{yu2018deepsolar,kausika2021geoai} or \cite{mayer20223d} leveraged convolutional neural networks (CNNs) and high-resolution overhead imagery to map solar installations over large areas; respectively the continental United States, the Netherlands and the German state of North-Rhine Westphalia. 

However, these mapping algorithms lack accountability since it is impossible to assess their accuracy while in deployment\cite{de2020monitoring}. This lack of accountability limits the practical use of these registries, as deep neural networks are sensitive to domain shift \cite{torralba2011unbiased}, which can lead to unpredictable performance drop under conditions that differ from those encountered in the training dataset. Currently, there exists no way to automatically and systematically assess the accuracy of PV mapping systems, although their sensitivity to domain shift is documented \cite{wang2017poor}. So far, the only attempts to assess the accuracy outside the training dataset have been to rely on manual verification \cite{kausika2021geoai,malof2019mapping}. 

Our main contribution is to propose an unsupervised and scalable way to monitor the accuracy of deep learning-based mapping pipelines over their whole {\it mapping area}. The {\it mapping area} covers the domain beyond the test set on which the model is deployed. We call the accuracy computed over the mapping area the {\it downstream task accuracy (DTA)}. We express DTA in an interpretable way for operators. To derive this measurement, we use the most detailed available information on distributed PV installations and compare it to the outputs of our model. We provide an unsupervised and interpretable way of assessing the accuracy of PV mapping algorithms. 

Additionally, we contribute to the ongoing effort to map distributed PV installations' tilt, localization, and installed capacity over more than 50,000 square kilometers in France, currently the largest PV registry with this level of detail.

\section{Related work}\label{sec:literature}

Scholars proposed numerous approaches over the last few years to map solar arrays on overhead imagery. \cite{de2020using} provides a complete overview of the works in this field. Early works relied on hand-crafted features and classification algorithms to identify PV panels on aerial images. The advent of deep-learning \cite{lecun2015deep} enabled large-scale mapping of PV panels using semantic segmentation. 

The DeepSolar project \cite{yu2018deepsolar} was a significant milestone as a deep learning-based pipeline was used to detect PV installations and estimate their surface area for the first time. The method relies on a two-step pipeline: images are classified, and if an image contains an array, it is passed to a segmentation model to identify the polygons corresponding to the PV installation. With this method, the authors achieved a precision of 93.1\% (recall: 88.5\%) in residential and a precision of 93.7\% (recall: 90.5\%) in nonresidential areas. 

DeepSolar triggered efforts to construct CNN-based pipelines to map installations over large areas \cite{mayer20223d,kausika2021geoai}. However, as pointed out by \cite{de2020monitoring}, pipelines developed over one territory cannot be straightforwardly applied to another. Moreover, the data extracted with automatic pipelines is not accountable enough to be used in official statistics. As a first step towards a better assessment of the accuracy of deep learning-based PV pipelines over their mapping area, \cite{kausika2021geoai} leveraged manual annotators to compute the precision and recall in places that were not in the training dataset. However, this method is labor-intensive and cannot be scaled up.

\cite{de2020monitoring} highlight two main research directions in PV detection. First, improving the reported data's accountability over the mapping area is necessary. To improve accountability, one needs to assess the accuracy of the pipeline's output in the mapping area. Second, one needs to mitigate domain shift as it degrades the accuracy of deep-learning mapping algorithms. In this work, we address the first question by introducing a scalable and unsupervised approach to assess the accuracy of a remote PV mapping algorithm over its mapping area.

\section{Data}

\subsection{Training data}\label{sec:training_data}

We train our classification and segmentation models on a new dataset called BDAPPV ({\it Base de données d'apprentissage profond photovoltaïque}). This dataset contains labeled thumbnails of PV panels. These PV panels come from a PV database maintained by the non-profit association of small owners of PV panels "Asso BDPV" ({\it Base de données photovoltaïque}). Table \ref{tab:statistics_bdappv} summarizes the characteristics of our training dataset. Our training dataset is nearly balanced, and we provide information on the PV systems' metadata, such as the tilt, azimuth, installed capacity, and other technical characteristics. A crowdsourcing campaign enabled us to gather this training data, which we release in \cite{kasmi2022crowdsourced}.

\begin{table}[h]
\centering
\caption{Training dataset characteristics. Column "Total number of samples" indicates the number of samples in the dataset (prior to augmentation), and column "positive samples" indicates the number of samples depicting a PV panel.}\label{tab:statistics_bdappv}
\begin{tabular}{| l | c | c |}
\hline
Dataset &  Total number of samples & Positive samples (share in \%)\\
\hline
Train &  12127 & 5445 (44.90) \\
Validation & 1732  & 755 (43.59) \\
Test & 3466  & 1485 (42.84)\\
\hline
Total &17325 &7685 (44.36)\\
\hline
\end{tabular}
\end{table}

\subsection{Geographical data for PV systems mapping}

We feed our classification and segmentation models into the PV mapping pipeline described in section \ref{sec:registry_pipeline}. We then deployed this pipeline over a large deployment area of 9 French {\it départements}, representing an area of more than 50 000 square kilometers. Our PV mapping pipeline requires the following data inputs:

\subsubsection{Orthoimagery} We do image classification and segmentation on RGB orthoimagery. These images are provided by the {\it Institut Géograhpique National} (IGN) and are freely accessible online. This dataset is called BD ORTHO. The ground sampling distance of these images is 20 cm/pixel. These images cover all French {\it départements}. For our study, we downloaded the image bundles of 9 French {\it départements}, located in the North, West, South, and East of France, covering approximately 10\% of the French metropolitan territory. Images of a given {\it départements} are updated every three years. In the worst case, we map the installations as they were three years ago. Since we know the overall installed capacity for a {\it département}, and assuming that the installations' characteristics are comparable over three years, the relatively low revisit rate is not problematic: we construct a representative sample of the current installed capacity. 

\subsubsection{Topographic data} We use topographic data, also provided by the IGN, under an open license. This dataset is called BD TOPO. This data contains the geographic footprints of all buildings registered in France. This dataset's main aim is to filter and merge detections made on the same rooftop.

\subsubsection{Distributed PV characteristics} At the characteristics extraction stage of our PV registry pipeline (see figure \ref{fig:flowchart}), we use the PV characteristics gathered in the BDPV database to calibrate the characteristics extraction module of our pipeline. These characteristics include the tilt, azimuth, surface, and installed capacity of PV installations. This database contains information on about 17000 installations in France, less than 5\% of the total estimated count of distributed PV installations. Moreover, as self-reported, the installations' geographical localization is not representative of the actual geographical localization of the distributed PV installations in France. 

\subsection{Monitoring data}

\subsubsection{National installations registry} Our data source for accuracy assessment is the {\it Registre national d'installations} (RNI) \cite{odre2022rni}. The French administration compiles the RNI every year. The RNI contains the total number of distributed PV installations and the aggregated installed capacity for each city. It does not contain the individual PV systems' characteristics. 

\section{Methods}

\subsection{Automated registry pipeline}\label{sec:registry_pipeline}

\begin{figure}[h]
\centering
\includegraphics[width=0.6\textwidth]{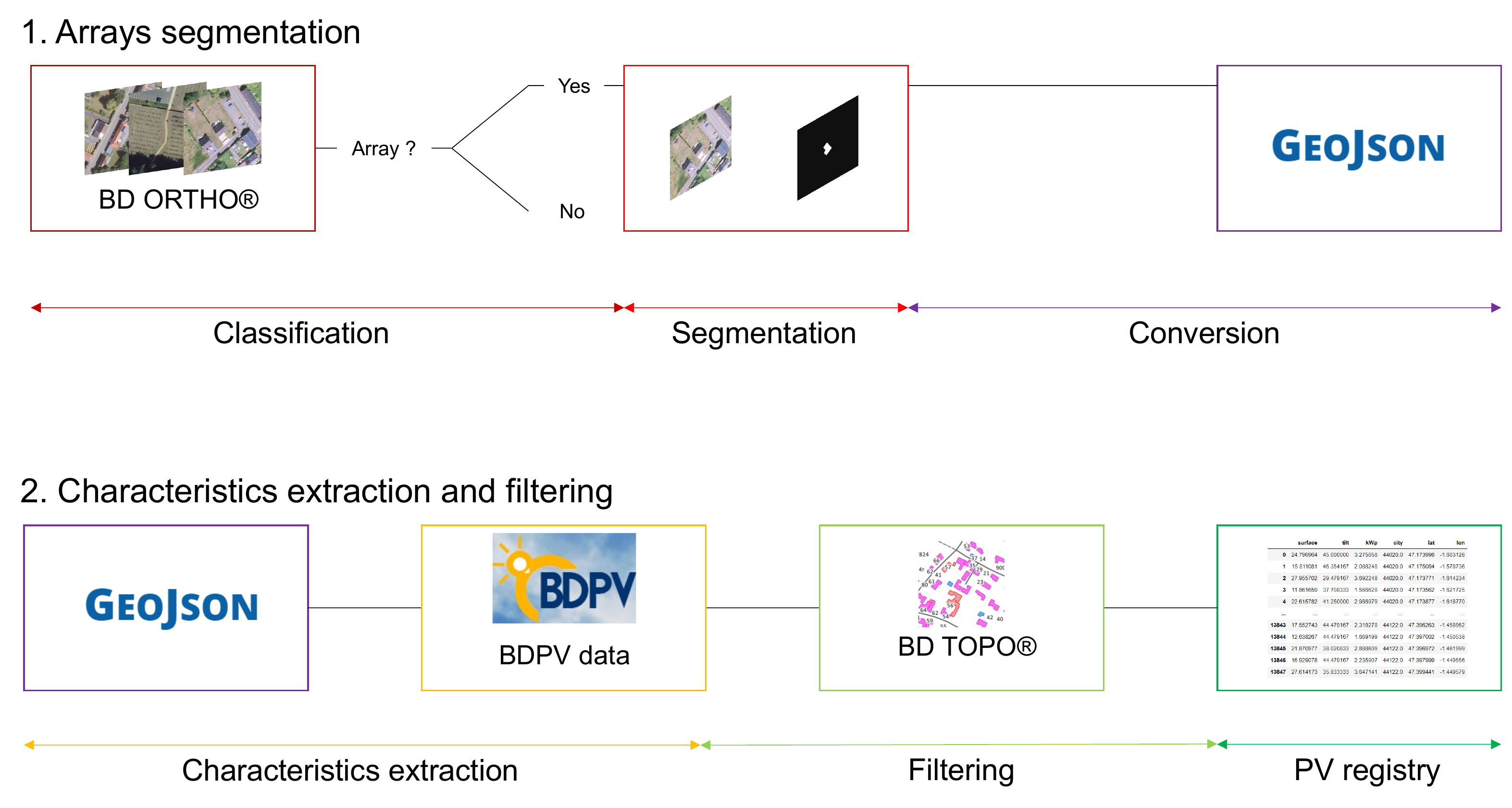}
\caption{Automated PV characteristics extraction pipeline} \label{fig:flowchart}
\end{figure}

We build on \cite{mayer20223d} to construct an automated PV registry pipeline. This pipeline takes as input orthoimages and topological data and outputs PV panels' characteristics (surface, installed capacity, tilt, and azimuth angles) and localization. Figure \ref{fig:flowchart} summarizes our approach. 

\subsubsection{Classification and segmentation} 

We fine-tuned our classification (Inception-v3) and segmentation (DeepLab-v3) models on our training dataset BDAPPV (see section \ref{sec:training_data}). During inference, we classify all images and segment only images on which a PV panel is detected. Segmentation masks are then converted into polygons. We use the weights of \cite{mayer20223d} as initialization for our fine-tuned models. Training details are provided in section \ref{sec:training_deeplearning}.

\subsubsection{Characteristics extraction} 

The output of the segmentation stage is a file with all PV installations' polygons detected in a given {\it département}. Based on this polygon, we extract the following characteristics: location, projected surface area, tilt angle, surface area, and installed capacity. The computation of the location and projected surface area are straightforward. Like previous works \cite{mayer20223d}, we derive the installed capacity from the surface area and the tilt angle using a linear fit. The regression coefficient captures the efficiency of the PV panels. 

\begin{figure}[h]
\centering
\includegraphics[width=0.6\textwidth]{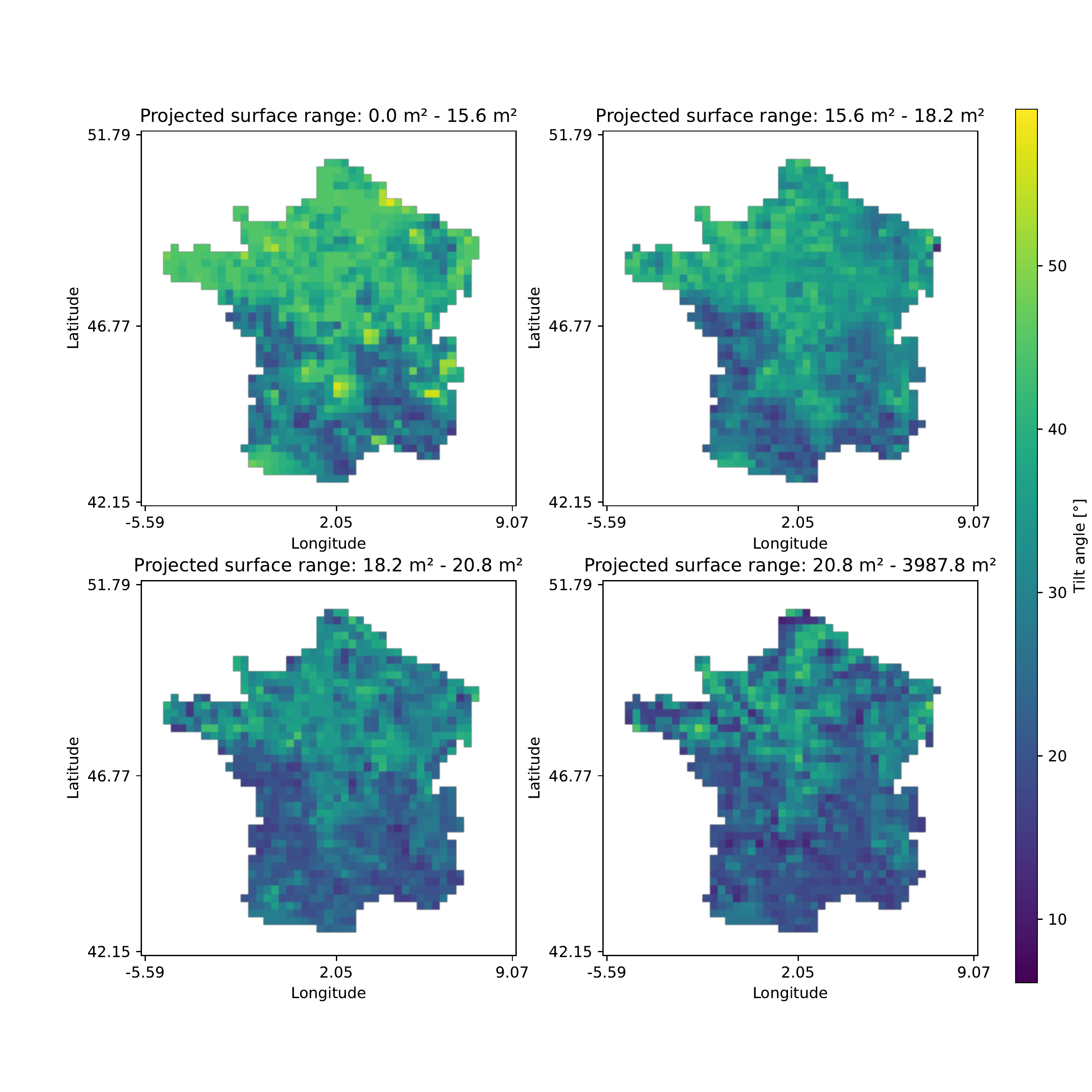}
\caption{Look-up table (LUT) to infer the tilt angle of the PV installations based on its localization and projected surface area. The LUT gives the average tilt over each square, computed based on the BDPV dataset.} \label{fig:lut}
\end{figure}

We computed a look-up table based on the BDPV metadata database \cite{kasmi2022crowdsourced} to estimate the tilt angle. The BDPV database spreads across France. This LUT enables us to impute a tilt angle based on the PV polygon's projected surface and localization. We defined four projected surface clusters based on their statistical occurrence in the BDPV database. Then, for each surface cluster (see fig. \ref{fig:lut}), we divide France into geographical squares, and for each square, we compute the average tilt angle. Figure \ref{fig:lut} depicts the look-up-table. We chose this approach as it is intuitive, computationally very efficient, and does not require surface models as in \cite{mayer20223d}. Our main aim is to capture the geographical variability of the tilt angle. As shown in figure \ref{fig:lut}, tilts are steeper in the north and for small installations. Besides, we also capture the fact that tilt angles are steeper for smaller installations \cite{saint2017probabilistic}.

\subsubsection{Post-processing} We use the BD TOPO to filter all polygons that are not on a building. We also filter installations that are either too small (we set the threshold at 1.7 square meters, the typical area of a single PV module) or too large. The upper threshold is 36 kWp as the RNI focuses on installations with an installed capacity lower than 36 kWp.

\subsection{Training procedure}\label{sec:training_deeplearning}

\subsubsection{Augmentations} During training, images are flipped vertically and rotated 90 degrees clockwise and counter-clockwise. We avoid rotations that lead to having arrays pointing north (i.e., upwards). Input images have a resolution of 400$\times$400, so we randomly crop the images to generate input images with the correct resolution of 299$\times$299 pixels. Finally, images are also normalized with the standard ImageNet values (i.e., a mean of $(0.485, 0.456, 0.406)$ and a standard deviation of $(0.229, 0.224, 0.225)$). No augmentation, but the normalization is applied to the validation and testing sets. 

\subsubsection{Classification branch} We retrain all the layers of the model. We train the model for 25 epochs. We pick the model that achieved the lowest validation accuracy after the end of the training. We evaluate the final performance after threshold fine-tuning on the testing dataset. We use the binary cross entropy loss (BCE, without weighting) and Adam optimizer with a learning rate of 0.0001. We used a batch size of 128.

\subsubsection{Segmentation branch} The model is trained for 25 epochs. As for classification, we pick the model that achieved the lowest validation accuracy after the end of the training. We also use BCE loss with a learning rate of 0.0001 and Adam optimizer. The batch size is 64, spread across multiple GPUs. 

\subsection{Downstream task accuracy metrics}\label{sec:continuous_accuracy}

The RNI provides city-wise the total number of installations and the aggregated installed capacity. To use it as a reference, we aggregate our detections city-wise. To the best of our knowledge, this is the first work to introduce a systematic assessment of the accuracy in terms of installed capacity over the whole deployment area. \cite{yu2018deepsolar} assessed the accuracy of the estimation of the surface of the PV panels on the test set only. \cite{mayer20223d} measured the accuracy of the installed capacity estimation per installation but only in cities for which reference data is available. 

\subsubsection{Definition} The {\it downstream task accuracy (DTA)} is a set of metrics that measures the accuracy of a model over the whole mapping area, where no labeled data is available. We express de DTA in interpretable terms for operators. In our case, we compute the DTA in terms of installed capacity using the models' outputs and the aggregated information of the RNI. The DTA requires no additional human labor as we automatically compute it.

\subsubsection{DTA metrics} Using the RNI dataset and the aggregations from our pipeline, we define three metrics to ensure that the model estimates the installed capacity correctly. We denote $\hat{C_i}$ our estimation of the installed capacity based on the aggregation of our detections in city $i$, $C_i$ the reference installed capacity from the RNI for city $i$. Similarly, $\hat{k}_i$ denotes our estimation of the number of installations in city $i$, and $k_i$ is the reference number of installations recorded in the RNI.

\begin{itemize}
\item The {\bf average percentage error (APE)} $\displaystyle{\frac{\vert C_i - \hat{C_i} \vert}{C_i}}$ and the {\bf mean APE (MAPE)} $\displaystyle{\frac{1}{n}\sum_{i=1}^n\frac{\vert C_i - \hat{C_i} \vert}{C_i}}$ computed over the whole {\it département}. The APE and MAPE are computed for the installed capacity $C_i$ in city $i$,
    \item The {\bf detection ratio $\displaystyle{\Delta := \frac{\hat{k_i}}{k_i}}$}, based on \cite{mayer20223d} computed at the city level and averaged over the {\it départements}. We compute this ratio for the number of installations $k_i$ in city $i$,
    \item The {\bf average installation percentage error (AIPE)} $\displaystyle{- \frac{C_i/k_i - \hat{C_i/k_i}}{C_i/k_i}}$ which is the APE computed for the {\it average} installation. By construction, a negative AIPE indicates that we underestimate the installations' size, and a positive AIPE indicates that we overestimate them. 
\end{itemize}

The MAPE ensures we do not overestimate or underestimate the overall installed capacity. The detection ratio ensures that we detect the correct number of installations, and the AIPE, which is a function of the installed capacity and the number of installations, ensures that, on average, we correctly estimate the size of the installations. 

\section{Results}

\subsection{Accuracy on the test set}

\subsubsection{Classification and segmentation accuracies}

Our fine-tuned model achieves competitive results compared to state-of-the-art models (see table \ref{tab:metrics}). For the classification branch, we reach an F1 score of 0.84. For the segmentation branch, we reach an Intersection-over-Union (IoU) of 0.86. Our aim is not to establish a new state-of-the-art (SOTA) for classification or segmentation but to see how current performance translates in terms of accuracy over the mapping area.

\begin{table}[h]
\centering
\caption{Classification and segmentation accuracy. The ground sampling distance (GSD) indicates how detailed the image is. The lower the GSD, the more detailed the image. PV panels cannot be detected on images that have a GSD greater than 30 cm/pixel (\cite{de2020monitoring}).}\label{tab:metrics}
\begin{tabular}{| c | c | c | c |}
\hline
 & {\bf Classification} &{\bf Segmentation}&\\
\hline
\hline
Work & F1-score & IoU & GSD (cm/pixel)\\
\hline
\cite{mayer20223d} & 0.87 & 0.74 & 10\\
\cite{malof2019mapping} & - & 0.67 & 30\\
\cite{zech2020predicting} & 0.82 & - & 10\\
\cite{parhar2021hyperionsolarnet} & {\bf 0.97} & 0.86 & 10\\
\hline
Ours & 0.84 & {\bf 0.86} & 20\\
\hline
\end{tabular}
\end{table}

\subsubsection{Comparison to existing accuracy metrics} As mentioned in section \ref{sec:continuous_accuracy}, there is no counterpart to our DTA metrics in the existing literature. However, \cite{yu2018deepsolar} computed the area estimation's mean relative error (MRE) and reported an error of 3.0\% in residential and 2.1\% in nonresidential areas. Our model achieves 0.44\% MRE on the test set. See \cite{yu2018deepsolar} for more details on the computation of the MRE. 

\cite{mayer20223d} computed the installation-wise accuracy over the cities of Borken, Unna, and Dortmund, Germany. They reported a mean average error (MAE) between 1.92 kWp and 24.84 kWp and a mean average percentage error (MAPE) between 27\% and 37\%. Our model achieves an MAE of 0.38 kWp and a MAPE of 11.59\% when comparing the estimated installed capacity with the reference recorded in the BDPV dataset.

\subsection{Downstream task accuracy on the mapping area}

We deploy our model on nine French {\it départements}: Nord (north), Loire-Atlantique (west), Hérault (south), and 6 out of 8 {\it départements} of the Rhône-Alpes region (southeast). This way, we ensure sufficient geographical diversity in landscapes, population densities, and architectural styles. The total area covered is 51858 square kilometers. It is currently the largest area mapped with this level of detail, as \cite{mayer20223d} covered an area of roughly 34000 square kilometers. 

We compute our DTA metrics over this mapping area using the RNI. To compute a baseline, we consider the test set as one city and compute the associated mean and median APE, detection ratio, and AIPE. We report the results in table \ref{tab:accuracy_large_scale}.

\subsubsection{Main results} As we can see in table \ref{tab:accuracy_large_scale}, our model performs worse on the mapping area than on the test set. The MAPE is about 30 percentage points higher on the mapping area than on the test set. This result is consistent with existing literature, and our accuracy tracking metrics enable us to quantify this decrease in accuracy. The MAPE is 47\%, and the mean ratio is slightly above 1. In detail, we can see that we tend to slightly overestimate (by 16\%) the size of the installations compared to the RNI. Performance for all metrics is relatively constant across all {\it départements} but the Nord.

\subsubsection{Effect of filtering on accuracy} As an example use case for the DTA, we discuss the effect of filtering using the topological data. Filtering by buildings decreases the number of detections and especially of small detections. Its effects on accuracy are ambiguous. On the one hand, it slightly biases the size of the installations (the ratio decreases and the AIPE increases). On the other hand, it can dramatically improve the detection accuracy in places where performance is low: in the {\it départements} Nord and Hérault, we can see that the post-processing improves all evaluation metrics by a large margin. 

\begin{table}[h]
\centering
\caption{Downstream task accuracy across the mapping area. Values in parentheses correspond to the results without filtering by buildings. The line "Test" considers the training dataset as one city. $k_i$ and $C_i$ denote the count of installations and the installed capacity, respectively. The hat indicates the estimation by our pipeline. }\label{tab:accuracy_large_scale}
\begin{tabular}{| c | c | c | c | c | c | c | c | c |}
\hline
{\it Dép.} & MAPE & med. APE & mean $\Delta$ & mean AIPE & $k_i$ & $\hat{k}_i$ & $C_i$ & $\hat{C}_i$\\
& [\%] & [\%] & [-] & [\%] & [-] & [-] & [kWp] & [kWp] \\
\hline
\hline
Test & 17.61 & - & 0.92 & -0.10 & 1485 & 1362 & 6473.8  & 5334,02\\
\hline
44 &  39.09 &  38.05 & 0.67 & 22.83  & 12683 & 6838 &  51955.17 & 34197.71\\
 & {\it (33.20)} & {\it (26.69)} & {\it (0.91)} & {\it (18.54)}&  & {\it (9325)} &  &  {\it (45206.58)}\\
\hline

69  & 31.99 & 28.91 & 0.83& 12.18 & 8944 &  6508 &  36500.6 &  31361.59\\
&{\it (39.45)} & {\it (23.29)} & {\it (1.26)} & {\it (10.33)} &  & {\it (9808)} & &  {\it (45433.94)}\\
\hline

59  & 130.06 & 88.13 & 2.23 &  61.16 &6453 & 9524 & 22083.8 & 52790.18\\
&{\it (224.22)} & {\it (168.29)} & {\it (3.22)} & {\it (41.56)} &  & {\it (15393)} & & {\it (73697.38)} \\
\hline

34  &26.80 & 17.78 & 1.01& 6.82 & 9199 & 8408 & 35398.41 & 39897.16 \\
&{\it (45.57)} & {\it (30.05)} & {\it (1.33)} & {\it (4.29)} &  & {\it (11445)}  && {\it (52841.79)}\\
\hline

 01  &35.90&35.35 & 0.77&6.18 & 4940 & 3654 & 18433.19 &  14659.39 \\
 &{\it (38.77)} &{\it ( 27.07)} & {\it (1.13)} & {\it (7.34)} &  & {\it (5259)} & & {\it (21371.97)}\\
 \hline
 
 38  &33.41&31.15 & 0.81&9.18 & 10672 & 7835 & 39691.43 & 32391.11 \\
 &{\it (31.29)} & {\it (21.80)} & {\it (1.07)} & {\it (6.20)} & & {\it (10680)} & & {\it (42617.43)} \\
 \hline
  
 42 &29.12&23.81 & 1.00&15.42 & 6892 & 5831 & 28594.08 & 27916.48 \\
 & {\it (46.30)} & {\it (28.66)} & {\it (1.41)} & {\it (11.06)}  && {\it (8384)} & & {\it (38222.44)} \\
 \hline
 
26 &30.46&23.51 & 0.92 &  3.74 & 5808 & 4933 & 28261.94 & 25833.96 \\
 & {\it (55.35)} & {\it (25.80 )}& {\it (1.43)} & {\it (6.15)} & & {\it (7121)} & & {\it (37645.72)} \\
 \hline

74 & 44.08 &  41.18 &  0.67 &  -4.61 & 7004 & 5287 & 32202.1 & 21760.4 \\
 & {\it (41.45)} & {\it (28.53)} & {\it (0.93)} & {\it (-6.10)}&  & {\it (6600)} &  & {\it (25930.6)} \\
\hline
Overall & 47.45 & 32.81 & 1.03 & 16.33 & 72595 & 58818 & 293120.70 & 280807.91\\
 & {\it (66.20)} & {\it (30.66)} & {\it (1.46)} & {\it (12.03)} & & {\it (84015)}&&{\it (382967.83)} \\
\hline
\end{tabular}
\end{table}

\subsection{Generalization to other areas}

The need for the French TSO for detailed information on distributed PV motivates this work. However, this need is ubiquitous in many European countries and beyond \cite{killinger2018search,shaker2015data}. It is possible to apply our accuracy assessment in these countries: for instance, the RNI can be substituted by the {\it Marktstammdatenregister (MaStR)} for Germany \cite{bundes2022markt} , the {\it Stamdataregister} for Denmark \cite{energi2022stamdata}. 

Besides, high-resolution orthoimagery is increasingly available across European countries due to the INSPIRE directive, which enforces open access to geographical information services (GIS) data \cite{ec2007inspire}. \cite{mayer20223d} used orthoimagery available over North Rhine Westphalia and \cite{kausika2021geoai} images available for the Netherlands. Although this case study focuses on France, the data required for replication is available throughout Europe.

\section{Conclusion and future work}

We built on existing literature to construct an automated pipeline for large-scale distributed photovoltaic (PV) mapping and characterization. We mapped about 50000 square kilometers over France, whose distributed PV installations are not well mapped, resulting in the largest with this level of detail.

Crucially, we address the need for a more transparent and systematic assessment of the accuracy beyond the test set through downstream task accuracy (DTA). DTA measures the accuracy in terms of installed capacity over the whole mapping area, i.e., the area over which the model is deployed. It is unsupervised and does not require additional human labor. DTA allows quantifying the decrease in accuracy when we shift from the test set to the mapping area. Such quantification paves the way for safer integration of deep learning-based pipelines for remote PV mapping.

We benchmarked state-of-the-art detection mapping pipelines with our newly introduced downstream task accuracy metrics. We showed that test set performance overestimates the accuracy of such pipelines in estimating the installed capacity of distributed PV installations by about 30 percentage points.

In future work, we will derive downstream task accuracy metrics for tilt and azimuth estimation. We also intend to pursue the model's deployment until eventually mapping the whole of metropolitan France. Finally, we shall discuss the benefit of more advanced classification and segmentation models on downstream task accuracy.

\bibliography{sample-ceur}

\end{document}